\documentclass[sigconf]{acmart}

\AtBeginDocument{%
  \providecommand\BibTeX{{%
    \normalfont B\kern-0.5em{\scshape i\kern-0.25em b}\kern-0.8em\TeX}}}

\setcopyright{acmcopyright}
\copyrightyear{2018}
\acmYear{2018}
\acmDOI{10.1145/1122445.1122456}

\acmConference[Woodstock '18]{Woodstock '18: ACM Symposium on Neural
  Gaze Detection}{June 03--05, 2018}{Woodstock, NY}
\acmBooktitle{Woodstock '18: ACM Symposium on Neural Gaze Detection,
  June 03--05, 2018, Woodstock, NY}
\acmPrice{15.00}
\acmISBN{978-1-4503-XXXX-X/18/06}

\usepackage{subcaption}
\usepackage{graphicx}
\usepackage{bm}
\usepackage{bbm}
\usepackage{url}
\usepackage{stmaryrd}
\usepackage{balance}
\usepackage{pgfplots}
\usepackage{pifont}
\usepackage{epsfig}
\usepackage{booktabs}% professional tables
\usepackage{pgffor}
\usepackage{tikz}
\usepackage{multirow}
\usepackage{url}
\usepackage[linesnumbered,ruled,vlined]{algorithm2e}

\makeatletter
\newcommand\xleftrightarrow[2][]{%
  \ext@arrow 9999{\longleftrightarrowfill@}{#1}{#2}}
\newcommand\longleftrightarrowfill@{%
  \arrowfill@\leftarrow\relbar\rightarrow}
\makeatother

\newcommand{\mb}{\mathbf}
\newcommand{\mc}{\mathcal}

\newtheorem{theo}{\textsc{Theorem}}

\newcommand{\gnlasso}{\textsc{Gnl}}
\newcommand{\gdu}{\textsc{Gdu}}
\newcommand{\linear}{\textsc{Linear}}
\newcommand{\lasso}{\textsc{Lasso}}
\newcommand{\gl}{\textsc{GLasso}}
\newcommand{\tvgl}{\textsc{TVGL}}

\newcommand{\lstm}{\textsc{LSTM}}
\newcommand{\rnn}{\textsc{RNN}}
\newcommand{\cnn}{\textsc{CNN}}
\newcommand{\gcn}{\textsc{GCN}}
\newcommand{\gnn}{\textsc{GNN}}
\newcommand{\gat}{\textsc{GAT}}
\newcommand{\gwn}{\textsc{Graph WaveNet}}
\newcommand{\mtgnn}{\textsc{MTGNN}}
\newcommand{\gman}{\textsc{GMAN}}
\newcommand{\dcrnn}{\textsc{DCRNN}}
\newcommand{\stgcn}{\textsc{STGCN}}
\newcommand{\mrabgcn}{\textsc{MRA-BGCN}}
\newcommand{\netinf}{\textsc{NetInf}}
\newcommand{\connie}{\textsc{ConNIe}}
\newcommand{\kernelcas}{\textsc{KernelCascade}}
\newcommand{\dsp}{\textsc{DSP}}

\setcopyright{rightsretained}

\begin{document}

% \acmConference{CIKM'17}{}{November 6--10, 2017, Singapore.}
% \acmPrice{15.00}
% \acmDOI{https://doi.org/10.1145/3132847.3132976}
% \acmISBN{ISBN 978-1-4503-4918-5/17/11} 
 
% Copyright

% \acmYear{2017} 
% \setcopyright{acmcopyright}
% \acmConference{CIKM'17}{}{November 6--10, 2017, Singapore.}\acmPrice{15.00}\acmDOI{10.1145/3132847.3133026}
% \acmISBN{978-1-4503-4918-5/17/11}

\fancyhead{}
%\settopmatter{printacmref=false, printfolios=false}

%----------------------------------------------------------------------------------------
\title{Graph Neural Lasso for Dynamic Network Regression}

\author{Yixin Chen, Lin Meng, Jiawei Zhang}
\email{yixin@ifmlab.org, lin@ifmlab.org, jiawei@ifmlab.org}
\affiliation{%
  \institution{IFM Lab, Florida State University, Tallahassee, FL 32311, USA}
}

%\author{Submitted For Blind Review} 

%\author{\IEEEauthorblockN{Jiawei Zhang$^\dagger$, Philip S. Yu$^\dagger$, Yuanhua Lv$^\ddagger$}
%\IEEEauthorblockA{$^\dagger$University of Illinois at Chicago, Chicago, IL, USA \\
%$^\ddagger$Microsoft Research, Redmond, WA, USA\\
%jzhan9@uic.edu, psyu@cs.uic.edu, yuanhual@microsoft.com}
%}

%\numberofauthors{4}
%\author{
%\alignauthor
%Jiawei~Zhang\\
%      \affaddr{University of Illinois at Chicago}\\
%      \affaddr{Chicago, IL, USA}\\
%       \email{jzhan9@uic.edu}
%\alignauthor
%Philip~S.~Yu\\
%      \affaddr{University of Illinois at Chicago, Chicago, IL, USA}\\
%       \email{psyu@cs.uic.edu}
%\alignauthor
%Limeng~Cui\\
%      \affaddr{University of Chinese Academy of Sciences, Beijing, China}\\
%       \email{lmcui932@163.com}
%\alignauthor
%Yuanhua~Lv\\
%       \affaddr{Microsoft, Sunnyvale, CA, USA}\\
%       \email{yuanhual@microsoft.com}
%}

%\author{Submitted for Blind Review\\}

\begin{abstract}
The regression of multiple inter-connected sequence data is a problem in various disciplines. Formally, we name the regression problem of multiple inter-connected data entities as the ``\textit{dynamic network regression}'' in this paper. Within the problem of stock forecasting or traffic speed prediction, we need to consider both the trends of the entities and the relationships among the entities. A majority of existing approaches can't capture that information together. Some of the approaches are proposed to deal with the sequence data, like LSTM. The others use the prior knowledge in a network to get a fixed graph structure and do prediction on some unknown entities, like GCN. To overcome the limitations in those methods, we propose a novel graph neural network, namely Graph Neural Lasso ({\gnlasso}), to deal with the dynamic network problem. {\gnlasso} extends the {\gdu} (gated diffusive unit) as the base neuron to capture the information behind the sequence. Rather than using a fixed graph structure, {\gnlasso} can learn the dynamic graph structure automatically. By adding the attention mechanism in {\gnlasso}, we can learn the dynamic relations among entities within each network snapshot. Combining these two parts, {\gnlasso} is able to model the dynamic network problem well. Experimental results provided on two networked sequence datasets, i.e., Nasdaq-100 and METR-LA, show that {\gnlasso} can address the network regression problem very well and is also very competitive among the existing approaches.

\end{abstract}


\maketitle

%-----------------------------------------------
\section{Introduction}\label{sec:introduction}

The network provides a general representation of many inter-connected data from various disciplines, e.g., human brain graphs~\cite{Bullmore_Brain_Psychology_11}, financial stock market~\cite{Namaki_Network_Physica_11} and offline traffic data analysis~\cite{asif2013spatiotemporal}. By modeling the data instances as the nodes and their potential relationships (or correlations) as the links, data collected from these areas can all be represented as networks. In many of the cases, the individual entities (i.e., the nodes) involved in the network are also associated with certain attributes whose values may change with time. The historical value changing records of the individual entities will lead to a set of time-series data points associated with the nodes. Due to the extensive links among the entities, the attribute changes of connected entities may display certain correlations; whereas, as the entity attribute value changes, the relationships among the instances should also evolve dynamically. 

%------------------------------------
\begin{figure}
    \centering
    \includegraphics[width=0.8\linewidth]{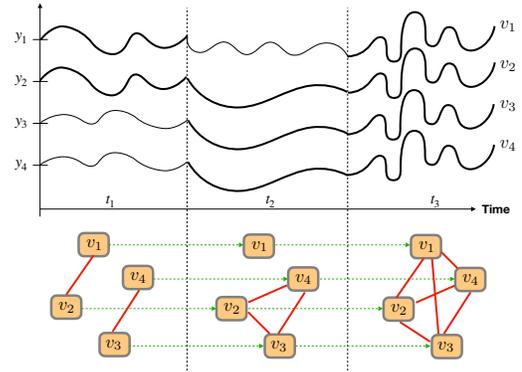}
    \caption{A toy example of dynamic network with four entities, $v_1$ to $v_4$: The curve shows the value of four stocks during time range $t_1$ to $t_3$. The x axis denotes the time and the y axis denotes the stock value. The graphs under the curve show how the correlations change over time.}
    \label{fig:example}
\end{figure}
%------------------------------------

Learning the attribute changing patterns of individual nodes and the evolutionary dynamics of the extensive links in the networks can both be important problems. For instance, with brain imaging techniques (e.g., fMRI, CT, or EEG)~\cite{Raichle_A_Trends_09}, the brain regional activity data can be represented as a dynamic network. Learning the brain activities and their correlations can provide crucial signals for illustrating some brain diseases and disorders, like Alzheimer's disease and cognitive impairment~\cite{Williams_Preventing_Evidence_10}. It is similar to the stock market network and the traffic network. By learning the stock price changing patterns and their relationships, financial quantitative analysts will be able to build models for more accurate stock price inference~\cite{Mitchell_Managerial_Business_00}. Meanwhile, based on the data captured by the sensors and the graph structure among the roads, the transportation agencies can predict the potential traffic congestion and better control the traffic~\cite{DBLP:conf/aaai/ZhengFW020}.

To better illustrate the changing patterns of data entities (i.e., nodes) in a dynamic network,  Figure~\ref{fig:example} shows a toy example. There are four inter-connected entities $v_1$ to $v_4$ with their time-changing attribute values. During time range $t_1$, $v_1$ is correlated to $v_2$, while $v_3$ is correlated to $v_4$, whose corresponding graph structure is also illustrated at the bottom. But such correlations are not stable, which will evolve as node attribute values change. For instance, during time range $t_2$, $v_2$, $v_3$ and $v_4$ are correlated, while $v_1$ is isolated. And during time range $t_3$, the correlation has changed again, and all four entities are correlated at this time. It's hard to capture a time-varying relationship and even the time they changed is unknown.

\noindent \textbf{Problem Studied}: In this paper, we will study the \textit{dynamic network regression} problem, which focuses on inferring both individual entities' changing attribute values and the dynamic relationships among the entities in the network data simultaneously. To resolve the problem, a novel graph neural network, namely \textit{graph neural lasso} ({\gnlasso}), will be proposed in this paper. To model the real-time changes of nodes in the network, {\gnlasso} extends \textit{gated diffusive unit} ({\gdu})~\cite{Zhang_Fake_CORR_18} to the regression scenario and uses it as the basic neuron unit. {\gnlasso} can effectively model the dynamic relationships among the nodes based on an \textit{attention mechanism}.

The {\gnlasso} model proposed in this paper is a brand new model and has clear distinctions with the existing approaches. Different from the regression models, e.g., {\lasso}~\cite{Tibshirani_Regression_Royal_94}, {\gl}~\cite{Friedman_Sparse_Bioinformatics_08} and {\tvgl}~\cite{Hallac_Network_KDD_17}, {\gnlasso} is a graph neural network model and can be extended to a deeper architecture for modeling much more complex input data. Meanwhile, compared with the existing graph neural networks for classifications, e.g., {\gcn}~\cite{Kipf_Semi_CORR_16} and {\gat}~\cite{Velickovic_Graph_ICLR_18}, {\gnlasso} is proposed for the regression task instead. What's more, different from existing graph neural networks for dynamic network modeling, e.g., {\dcrnn}~\cite{DBLP:conf/iclr/LiYS018}, {\gnlasso} doesn't need to take the network structure information as the input, which can be inferred by {\gnlasso} instead in the learning process. In addition, besides the \textit{neural gate} and \textit{attention mechanism}, {\gnlasso} also adds an $L_1$-norm regularization on the model variables, which can resolve the overfitting \cite{tibshirani1996regression} problem common in regression models. The main contributions of this work can be summarized as follows:

\begin{itemize}
    \item We introduce a new graph neural lasso model {\gnlasso} in this paper to address the dynamic network regression task. {\gnlasso} extends {\gdu}~\cite{Zhang_Fake_CORR_18} to the regression problem settings, which will be used as a base neuron for node state modeling.
    \item We propose to use attention mechanisms in {\gnlasso} to model the dynamic relationships among the entities. We use the output after every step to learn the coefficient. Then, we use the coefficient to get a compound of the outputs with related entities. This new vector contains information with related entities and updated relationships.
    \item We add $L_1$-norm on the model to address the overfitting problem, which effectively help the model to learn sparse variable matrices.
    \item To illustrate the effectiveness of our method {\gnlasso}, extensive experiments have been done on two real-world dynamic network datasets, i.e., Nasdaq-100 and METR-LA. The experimental results demonstrate that {\gnlasso} is competitive among existing graphical regression models.
\end{itemize}

The rest of the paper is organized as follows. We discuss several existing approaches that can deal with the dynamic network problem in Section~\ref{sec:relatedwork}. Then, we introduce the {\gnlasso} in detail in Section~\ref{sec:method}. Also, we analyze the convergence of loss function with $L_1$-norm term in Section~\ref{sec:model_learning}. At last, we evaluate our model in Section~\ref{sec:experiment} and give the conclusion in Section~\ref{sec:conclusion}.

%-----------------------------------------------
\section{Related Work} \label{sec:relatedwork}
Our work relates to deep learning for time series forecasting, network inference, and graph neural networks.

\subsection{Deep Learning for Time Series Forecasting}
Deep learning for time series forecasting has been studied for many years. One type of extensive works on the prediction time series is the Recurrent Neural Network ({\rnn}) based models. Generally, many researchers developed different kinds of {\rnn} cells. Among them,  
{\lstm}~\cite{Hochreiter_Long_Neural_97} is one widely used variant of recurrent neural networks. {\rnn}s are networks with loops in them, allowing information to persist. Based on the simple cell of {\rnn}, {\lstm} adds a forget gate to discard some information at every cell state. The vanilla {\rnn} may have a long-term dependency problem since it can not capture the information with a large gap. Since {\lstm} has a more complex structure, it can do prediction on the sequence data with much more complicated models. Besides LSTM, GRU provides a simplified architecture for modeling long and short term memory instead \cite{DBLP:journals/corr/ChungGCB14}. It doesn't have a memory cell and uses fewer parameters. Recent works~\cite{rodrigues2019combining,gensler2016deep,DBLP:conf/iclr/LiYS018,DBLP:journals/tsg/KongDJHXZ19,DBLP:conf/cvpr/0005LCZG18} draw another research line for time series forecasting. In these works, hybrid models are proposed. For example,  in~\cite{rodrigues2019combining}, it combines time series and textual data to predict the taxi demand. However, simply considering the changing patterns of individual entities ignores the changing dependency among these entities. Learning the dependency among entities could help us get a better result.

\subsection{Network Inference}
Network inference is another related topic. Constructing the graph can help us better describe, analyze, and visualize the data~\cite{DBLP:journals/spm/DongTRF19}, widely used in domains like transportation networks, social networks, and brain networks. Graphical Lasso~\cite{Friedman_Sparse_Bioinformatics_08} ({\gl}) calculates a covariance matrix, which can give the correlations between a pair of entities. Note that the {\gl} can only calculate one covariance matrix based on the whole data. To better model a changing graph, time-varying graphical lasso~\cite{Hallac_Network_KDD_17} ({\tvgl}) uses temporal information to adjust the covariance matrix over time. These methods use statistical models to calculate the correlations. Some recent works~\cite{DBLP:journals/tkdd/Gomez-RodriguezLK12,DBLP:conf/nips/MyersL10,DBLP:conf/nips/DuSSY12,DBLP:journals/tsp/SandryhailaM13} also focus on the physically motivated models. {\netinf}~\cite{DBLP:journals/tkdd/Gomez-RodriguezLK12}, {\connie}~\cite{DBLP:conf/nips/MyersL10}, {\kernelcas}~\cite{DBLP:conf/nips/DuSSY12} are proposed to observe the paths of the diffusion and infer the graphs. While {\dsp}~\cite{DBLP:journals/tsp/SandryhailaM13} model the graphs by learning the signal representation.

\subsection{Graph Neural Networks}
Graph Neural Networks~({\gnn}) are one of the hottest topics in recent years. Graph Convolutional Network ({\gcn}) is one of the most popular models used in many areas such as computer vision~\cite{yao2018exploring,ling2019fast} and neural language processing~\cite{marcheggiani2017encoding,yao2019graph}. Whereas Graph Attention Networks learns the edge weights for all one-hop neighboring nodes, which enable us to capture the dependency of nodes. Recently, various kinds of graph neural networks have been proposed to deal with the regression problem on graphs. Those graph neural networks are designed to solve dynamic network problems, such as traffic prediction~\cite{wu2019graph,DBLP:journals/corr/abs-2005-11650,DBLP:conf/aaai/ZhengFW020,DBLP:conf/iclr/LiYS018,DBLP:conf/aaai/ChenCXCGF20,DBLP:conf/ijcai/YuYZ18}. They are all based on {\rnn} or {\cnn}. {\dcrnn}~\cite{DBLP:conf/iclr/LiYS018} combines the diffusion convolution layer and recurrent layer with an encoder-decoder architecture. {\stgcn}~\cite{DBLP:conf/ijcai/YuYZ18}, Graph Wavenet~\cite{wu2019graph} and {\mtgnn}~\cite{DBLP:journals/corr/abs-2005-11650} use graph convolution layer to capture the spatial information and temporal convolution layer to capture the temporal information. There are two other approaches, {\gman}~\cite{DBLP:conf/aaai/ZhengFW020} and {\mrabgcn}~\cite{DBLP:conf/aaai/ChenCXCGF20}, which include the attention mechanism to aggregate the information in different neighbors and model the relationships.

% In this paper, we utilize the graph dependency to handle the regression problem by GNL. It is a graph neural network with a special cell proposed in this paper. At each cell, it aggregates the information from the last cell of neighbor nodes. Then, with input data, it uses forget gate, evolves gate, and selection gate to gain a state information vector. By the design of the cell, GNL can infer the dynamic relationships among the entities and capture the sequence dependency to do regression.

%-----------------------------------------------
\section{Graph Neural Lasso}\label{sec:method}

In this section, we will first introduce the notations used in this paper, and then describe the extended {\gdu} neuron for the dynamic network regression problem. After that, we will introduce the attentive aggregation operator for neighbor information integration and the {\gnlasso} model architecture as well as its learning process.

%------------------------------------------------------------------------

\subsection{Notation}\label{subsec:notation}

In the sequel of this paper, we will use the lower case letters (e.g., $x$) to represent scalars, lower case bold letters (e.g., $\mb{x}$) to denote column vectors, bold-face upper case letters (e.g., $\mb{X}$) to denote matrices, and upper case calligraphic letters (e.g., $\mc{X}$) to denote sets or high-order tensors. Given a matrix $\mb{X}$, we denote $\mb{X}(i,:)$ and $\mb{X}(:,j)$ as its $i_{th}$ row and $j_{th}$ column, respectively. The ($i_{th}$, $j_{th}$) entry of matrix $\mb{X}$ can be denoted as either $\mb{X}(i,j)$ or $\mb{X}_{i,j}$, which will be used interchangeably. We use $\mb{X}^\top$ and $\mb{x}^\top$ to represent the transpose of matrix $\mb{X}$ and vector $\mb{x}$. For vector $\mb{x}$, we represent its $L_p$-norm as $\left\| \mb{x} \right\|_p = (\sum_i |\mb{x}(i)|^p)^{\frac{1}{p}}$. The Frobenius-norm of matrix $\mb{X}$ is represented as $\left\| \mb{X} \right\|_F = (\sum_{i,j} |\mb{X}(i,j)|^2)^{\frac{1}{2}}$. The element-wise product of vectors $\mb{x}$ and $\mb{y}$ of the same dimension is represented as $\mb{x} \otimes \mb{y}$, whose concatenation is represented as $\mb{x} \sqcup \mb{y}$.

%------------------------------------------------------------------------

%------------------------------------------------------------------------

\subsection{Gated Diffusive Unit}

The {\gdu} neuron was initially introduced for modeling the diverse connections in heterogeneous information networks \cite{Zhang_Fake_CORR_18}, which can accept multiple inputs from the neighbor nodes in networks. In this part, we will extend it to the dynamic network regression problem settings, and use it to model both the network snapshot internal connections and the temporal dependency relationships between sequential network snapshots for the nodes.

%------------------------------------------------------------------------

%------------------------------------
\begin{figure}
    \centering
        \includegraphics[width=\linewidth]{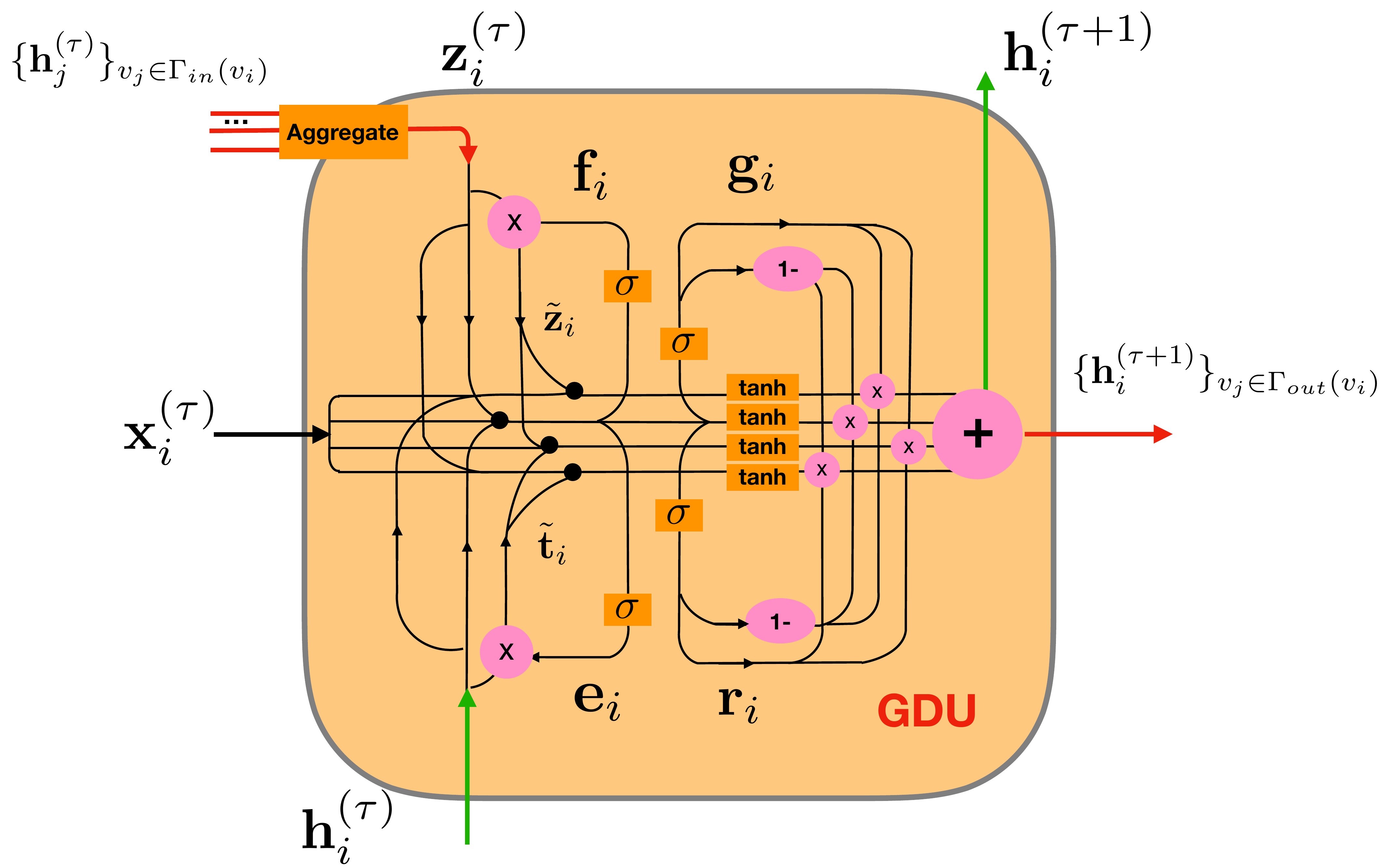}
        \caption{The detailed architecture of the {\gdu} neuron of node $v_i$ at timestamp $\tau$.}
        \label{fig:unit}
\end{figure}
%------------------------------------

\begin{algorithm}
 \SetAlgoLined
 \KwIn{$\{\mb{x}_i^{(\tau)}\}$: input data, $\mb{A}$: initial adjency matrix, $\{\mb{h}_i^{(0)}\}$: randomly initialized hidden state vector, $\beta$: the coefficient for $L_1$-norm term,  
}

 \KwResult{$\mb{h}^{\tau+1}_i$: vector representation}
 \While{$k$ < Epochs}{
  \While{$\tau$ < Length of input timestep}{
  \While{$i$ < Number of entities}{
   $\mb{z}_i^{(\tau)} = \mbox{Aggregate} \left(\{\mb{h}_j^{(\tau)}\}_{v_j \in \Gamma_{in}(v_i)} \right)$;
   
   $\mb{h}_i^{(\tau+1)} = \gdu(\mb{x}_i^{(\tau)}, \mb{h}_i^{(\tau)}, \mb{z}_i^{(\tau)})$;
   $i = i+1$;
  }
  $\tau = \tau+1$;
}
$\mathcal{L} = \min_{\boldsymbol{\Theta}} \ell(\boldsymbol{\Theta}) + \beta \cdot \left\| \boldsymbol{\Theta} \right\|_1$;

Update the model variable via backpropagation;

$k = k+1$;
 }
 \caption{{\gnlasso} framework} \label{alg:algo}
\end{algorithm}

%----------------------------------

Formally, given the time series data of connected entities, we can represent them as a dynamic network set $\mc{G} = \{G^{(1)}, G^{(2)}, \cdots, G^{(t)}\}$, where $t$ denotes the maximum timestamp. For each network $G^{(\tau)} \in \mc{G}$, it can be denoted as $G^{(\tau)} = (\mc{V}^{(\tau)}, \mc{E}^{(\tau)})$ involving the node set $\mc{V}^{(\tau)}$ and link set $\mc{E}^{(\tau)}$, respectively. Given a node $v_i$ in network $G^{(\tau)}$, we can represent its in-neighbors and out-neighbors as sets $\Gamma_{in}(v_i) = \{v_j | v_j \in \mc{V}^{(\tau)} \land (v_j, v_i ) \in \mc{E}^{(\tau)}\}$ and $\Gamma_{out}(v_i) = \{v_j | v_j \in \mc{V}^{(\tau)} \land (v_i, v_j ) \in \mc{E}^{(\tau)}\}$. Here, we need to add a remark that the link direction denotes the influences among the nodes. If the influences in the studied networks are bi-directional, we can have $\Gamma_{in}(v_i) = \Gamma_{out}(v_i)$ by default.

For node $v_i$ in network $G^{(\tau)}$ of the $\tau_{th}$ timestamp, we can denote the input attribute values of $v_i$ as an input feature vector $\mb{x}_i^{(\tau)} \in \mathbbm{R}^{d_x}$ ($d_x$ denotes the input raw feature dimension). {\gdu} maintains a hidden state vector for each node, and the vector of node $v_i$ at timestamp $\tau$ can be denoted as $\mb{h}_i^{(\tau)} \in \mathbbm{R}^{d_h}$ ($d_h$ denotes the hidden state vector dimension). As illustrated in Figure~\ref{fig:unit}, besides the feature vector $\mb{x}_i^{(\tau)}$ and hidden state vector $\mb{h}_i^{(\tau)}$ inputs, the {\gdu} neuron of $v_i$ will also accept the inputs from $v_i$'s input neighbor nodes, i.e., $\{\mb{h}_j^{(\tau)}\}_{v_j \in \Gamma_{in}(v_i)}$, which will be integrated via certain aggregation operators:

\begin{equation}\label{equ:aggregate}
\mb{z}_i^{(\tau)} = \mbox{Aggregate} \left(\{\mb{h}_j^{(\tau)}\}_{v_j \in \Gamma_{in}(v_i)} \right).
\end{equation}
The $\mbox{Aggregate}(\cdot)$ operator used in {\gnlasso} will be introduced in detail in the next subsection.

In dynamic network settings, the node states will change with time dramatically, and inputs from the previous network snapshots will become obsolete quickly. To resolve such a problem, besides the attention mechanism to be introduced later, {\gdu} introduces several gates for the neural state adjustment. Formally, for the aggregated inputs from the neighbor nodes, certain information in vector $\mb{z}_i^{(\tau)}$ can be useless for the state update of node $v_i$. To remove such useless information, {\gdu} defines a \textit{forget gate} $\mb{f}_i$ to adjust its representations as follows:
\begin{equation}
\tilde{\mb{z}}_{i}^{(\tau)} = \mb{f}_i \otimes \mb{z}_i^{(\tau)}, \mbox{ where  }\mb{f}_i = \sigma \left( \mb{W}_f \left[\mb{x}_i^{(\tau)} \sqcup \mb{z}_i^{(\tau)} \sqcup \mb{h}_i^{(\tau)} \right] \right),
\end{equation}
where $\sigma(\cdot)$ denotes sigmoid activation function and $\mb{W}_f$ is the involved variable. In the above equation, the super-script $^{(\tau)}$ of the gate is omitted for simpler representations. Meanwhile, {\gdu} also introduces a similar \textit{evolve gate} $\mb{e}_i$ for adjusting the hidden state vector input of $v_i$:
\begin{equation}
\tilde{\mb{h}}_i^{(\tau)} = \mb{e}_i \otimes \mb{h}_i^{(\tau)}, \mbox{ where  }\mb{e}_i = \sigma \left( \mb{W}_e \left[\mb{x}_i^{(\tau)} \sqcup \mb{z}_i^{(\tau)} \sqcup \mb{h}_i^{(\tau)} \right] \right).
\end{equation}
Neuron {\gdu} effectively integrates the useful information from $\mb{x}_i^{(\tau)}$, $\tilde{\mb{z}}_{i}$ and $\tilde{\mb{h}}_i$ to define the updated hidden state of node $v_i$. Instead of simple summation, integration of such information is achieved via two \textit{selection gates} $\mb{g}_i$ and $\mb{r}_i$ indicated as follows: \begingroup\makeatletter\def\f@size{8}\check@mathfonts

\begin{equation}\label{equ:gdu}
\hspace{-5pt}
\begin{aligned}
\mb{h}_i^{(\tau+1)}  \hspace{-2pt} &= \mb{g}_i \otimes \mb{r}_i \otimes \tanh \left(\mb{W}_u  \hspace{-2pt} \left[\mb{x}_i^{(\tau)}  \hspace{-2pt} \sqcup \tilde{\mb{z}}_i^{(\tau)}  \hspace{-2pt} \sqcup \tilde{\mb{h}}_i^{(\tau)} \right] \right) \\
&+ (\mb{1} - \mb{g}_i) \otimes \mb{r}_i \otimes \tanh \left(\mb{W}_u  \hspace{-2pt} \left[\mb{x}_i^{(\tau)}  \hspace{-2pt} \sqcup \mb{z}_i^{(\tau)}  \hspace{-2pt} \sqcup \tilde{\mb{h}}_i^{(\tau)} \right] \right)\\
&+ \mb{g}_i \otimes (\mb{1} - \mb{r}_i) \otimes \tanh \left(\mb{W}_u  \hspace{-2pt} \left[\mb{x}_i^{(\tau)}  \hspace{-2pt} \sqcup \tilde{\mb{z}}_i^{(\tau)}  \hspace{-2pt} \sqcup \mb{h}_i^{(\tau)} \right] \right)\\
&+ (\mb{1} - \mb{g}_i) \otimes (\mb{1} - \mb{r}_i) \otimes \tanh \left(\mb{W}_u  \hspace{-2pt} \left[\mb{x}_i^{(\tau)}  \hspace{-2pt} \sqcup \mb{z}_i^{(\tau)}  \hspace{-2pt} \sqcup \mb{h}_i^{(\tau)} \right] \right), \\
\hspace{-5pt}
&\mbox{where }
 \hspace{-2pt}
\begin{cases}
\mb{g}_i \hspace{-10pt} &= \hspace{-2pt} \sigma \left( \mb{W}_g  \hspace{-2pt} \left[\mb{x}_i^{(\tau)}  \hspace{-2pt} \sqcup \mb{z}_i^{(\tau)}  \hspace{-2pt} \sqcup \mb{h}_i^{(\tau)} \right] \right);\vspace{10pt}\\
\mb{r}_i \hspace{-10pt} &= \hspace{-2pt} \sigma \left( \mb{W}_r  \hspace{-2pt} \left[\mb{x}_i^{(\tau)}  \hspace{-2pt} \sqcup \mb{z}_i^{(\tau)}  \hspace{-2pt} \sqcup \mb{h}_i^{(\tau)} \right] \right).
\end{cases}
\end{aligned}
\end{equation}\endgroup

In the above equation, vector $\mb{1}$ denotes a vector filled with value $1$ of the same dimensions as the \textit{selection gate} vectors $\mb{g}_i$ and $\mb{r}_i$. Operator $\tanh(\cdot)$ denotes the hyperbolic tangent activation function and $\otimes$ denotes the entry-wise product as introduced in Section~\ref{subsec:notation}.

%------------------------------------------------------------------------

\subsection{Attentive Neighborhood Influence Aggregation Operator}

In this part, we will define the $\mbox{Aggregate}(\cdot)$ operator used in Equ.~(\ref{equ:aggregate}) for node neighborhood influence integration. The {\gnlasso} model defines such an operator based on an attention mechanism. Formally, given the node $v_i$ and its in-neighbor set $\Gamma_{in}(v_i)$, for any node $v_j \in \Gamma_{in}(v_i)$, {\gnlasso} quantifies the influence coefficient of $v_j$ on $v_i$ based on their hidden state vectors $\mb{h}_j^{(\tau)}$ and $\mb{h}_i^{(\tau)}$ as follows: \begingroup\makeatletter\def\f@size{9.5}\check@mathfonts
\begin{equation}\hspace{-5pt}
\begin{aligned}
\alpha_{j, i}^{(\tau)} &= \mbox{AttInf}(e_{j,i}^{(\tau)}) = \frac{\exp(e_{j,i}^{(\tau)})}{\sum_{v_k \in \Gamma_{out}(v_j)} \exp(e^{(\tau)}_{j,k})} ,\\
&\mbox{where }
e_{j,i}^{(\tau)} = \mbox{Linear}( \mb{W}_a \mb{h}_j^{(\tau)} \sqcup \mb{W}_a \mb{h}_i^{(\tau)}; \mb{w}_a ).
\end{aligned}
\end{equation}\endgroup
In the above equation, operator $\mbox{Linear}(\cdot; \mb{w}_a)$ denotes a linear sum of the input vector parameterized by weight vector $\mb{w}_a$. According to \cite{Velickovic_Graph_ICLR_18}, out of the model learning concerns, the above influence coefficient term can be slightly changed by adding the LeakyReLU function into its definition. Formally, the final influence coefficient used in {\gnlasso} is represented as follows:\begingroup\makeatletter\def\f@size{8.0}\check@mathfonts
\begin{equation}
\begin{aligned}
\hspace{-5pt}
\alpha^{(\tau)}_{j, i}  &=   \mbox{\small AttInf}(\mb{h}_j^{(\tau)}, \mb{h}_i^{(\tau)}; \mb{W}_a, \mb{w}_a)\\
&=  \frac{\exp ( \mbox{\small LeakyReLU} ( \mbox{\small Linear}( \mb{W}_a \mb{h}_j^{(\tau)} \sqcup \mb{W}_a \mb{h}_i^{(\tau)}; \mb{w}_a ) ) )}{\sum_{v_k \in \Gamma_{out}(v_j)} \exp ( \mbox{\small LeakyReLU} ( \mbox{\small Linear}( \mb{W}_a \mb{h}_j^{(\tau)} \sqcup \mb{W}_a \mb{h}_k^{(\tau)}; \mb{w}_a ) ) )}.
\end{aligned}
\end{equation}\endgroup

Considering that in our problem setting the links in the dynamic networks are unknown and to be inferred, the above influence coefficient term $\alpha_{j, i}$ actually quantifies the existence probability of the influence link $(v_j, v_i)$, i.e., the inference results of the links. Furthermore, based on the influence coefficient, we can provide the concrete representation of Equ.~(\ref{equ:aggregate}) as follows:
\begin{equation}
\begin{aligned}
\mb{z}_i^{(\tau)} &= \mbox{Aggregate} \left(\{\mb{h}_j^{(\tau)}\}_{v_j \in \Gamma_{in}(v_i)} \right)\\
&= \sigma \left( \sum_{v_j \in \Gamma_{in}(v_i)} \alpha^{(\tau)}_{j,i} \mb{W}_a \mb{h}_j^{(\tau)} \right).
\end{aligned}
\end{equation}

%------------------------------------------------------------------------

\subsection{Graph Neural Lasso with Dynamic Attentions}

In this part, we will introduce the architecture of the {\gnlasso} model together with its learning settings. Formally, given the input dynamic network set $\mc{G} = \{G^{(1)}, G^{(2)}, \cdots, G^{(t)}\}$, {\gnlasso} shifts a window of size $\tau$ along the networks in the order of their timestamps. The network snapshots covered by the window, e.g., $G^{(k)}$, $G^{(k+1)}$, $\cdots$, $G^{(k+\tau-1)}$, will be taken as the model input of {\gnlasso} to infer the network $G^{(k+\tau)}$ in following timestamp (where $k, k+1, \cdots, k+\tau \in \{1, 2, \cdots, t\}$). According to the above descriptions, we can denote the inferred attribute values of all the nodes and their potential influence links in network $G^{(k+\tau)}$ as
\begin{equation}
\begin{aligned}
&\hat{\mb{x}}_i^{(\tau+1)} = \mbox{FC}( \mb{h}_{i}^{(\tau+1)}; \boldsymbol{\Theta} ), \forall v_i \in \mc{V}^{(\tau+1)}; \\
&\alpha^{(\tau)}_{j,i} = \mbox{AttInf}(\mb{h}_{j}^{(\tau+1)}, \mb{h}_{i}^{(\tau+1)}; \boldsymbol{\Theta}), \forall v_i, v_j \in \mc{V}^{(\tau+1)},
\end{aligned}
\end{equation}
In the above equation, term $\mb{h}_{i}^{(\tau+1)}$ is defined in Equ.~(\ref{equ:gdu}) and $\boldsymbol{\Theta}$ covers all the involved variables used in the {\gnlasso} model. By comparing the inferred node attribute values, e.g., $\hat{\mb{x}}_i^{(\tau+1)}$, against the ground truth values, e.g., ${\mb{x}}_i^{(\tau+1)}$, the quality of the inference results by {\gnlasso} can be effectively measured with some loss functions, e.g., mean square error:
\begin{equation}
\begin{aligned}
\ell(\boldsymbol{\Theta}) &= \frac{1}{|\mc{V}^{(\tau+1)}|} \sum_{v_i \in \mc{V}^{(\tau+1)}} \ell(v_i; \boldsymbol{\Theta})\\
&= \frac{1}{|\mc{V}^{(\tau+1)}|} \sum_{v_i \in \mc{V}^{(\tau+1)}} \left\| \hat{\mb{x}}_i^{(\tau+1)} - {\mb{x}}_i^{(\tau+1)}\right\|_2^2.
\end{aligned}
\end{equation}
In addition, similar to {\lasso}~\cite{tibshirani1996regression}, to avoid overfitting, {\gnlasso} proposes to add a regularization term in the objective function to maintain the sparsity of the variables. Formally, the final objective function of the {\gnlasso} model can be represented as follows:
\begin{equation}\label{equ:objective_function}
\min_{\boldsymbol{\Theta}} \ell(\boldsymbol{\Theta}) + \beta \cdot \left\| \boldsymbol{\Theta} \right\|_1,
\end{equation}
where term $\left\| \boldsymbol{\Theta} \right\|_1$ denotes the sum of the $L_1$-norm regularizer of all the involved variables in the model and $\beta$ is the hyper-parameter weight of the regularization term. The pseudo-code for learning the above model is also illustrated in Algorithm~\ref{alg:algo}

%-----------------------------------------------
\section{Model Learning and Analysis}\label{sec:model_learning}

Different from the existing neural networks, from the objective function in Equation~(\ref{equ:objective_function}), we observe that the regularization term is the $L_1$-norm of the variables, which is non-differentiable. To effectively learn the model, in this paper, we propose several different learning techniques to resolve such a problem, where effectiveness will be tested in the following Section~\ref{gradient} with experiments on real-world datasets.

\subsection{Frobenius Norm Approximation}

One approach to make the model learnable is to approximate the $L_1$-norm on the regularization term with several other differentiable norms instead. One common approach is to replace the $L_1$-norm with the F-norm instead. Previous works~\cite{yan2009comparison,melkumova2017comparing} show that $L_1$-norm and F-norm get similar results. Formally, to use the F-norm, we can rewrite the objective function as follows:
\begin{equation}
\min_{\mb{\Theta}} \ell(\mb{\Theta}) + \beta \cdot \left\| \mb{\Theta} \right\|_F ,
\end{equation}
where $\left\| \mb{\Theta} \right\|_F$ denotes the sum of the F-norm based regularization terms on the variables.

Such an objective function can be effectively learned with the back-propagation algorithm. To be more specific, the training process involves multiple epochs. In each epoch, the training data is segmented subject to a window size, where the former time-series values are used as the known features and the last value is used as the objective value to be inferred. The sampling process goes among all the nodes in the network. Such a process continues until either convergence or the training epochs have been finished. The learned model is also named as ``{\gnlasso}-F`` in this paper.

\subsection{Piecewise Derivative}

The $L_1$-norm term is non-differentiable only at $\boldsymbol{\Theta}=\boldsymbol{0}$, where $\boldsymbol{\Theta}$ is the element in $\mb{\Theta}$. Thus, another common approach~\cite{bach2010sparse,regularization2013} is to set the derivative of $L_1$-norm term at $\boldsymbol{\Theta}=\boldsymbol{0}$:
\begin{equation}
\frac{\partial ||\boldsymbol{\Theta}||_1}{\partial \boldsymbol{\Theta}} = \boldsymbol{0} \text{, where } \boldsymbol{\Theta}=\boldsymbol{0}.
\end{equation}
Then, we can follow the piecewise derivative function below for $L_1$-norm during backpropagation:
\begin{equation}
\frac{\partial ||\boldsymbol{\Theta}||_1}{\partial \boldsymbol{\Theta}} = \left\{ 
\begin{aligned}
\boldsymbol{-1}&, \boldsymbol{\Theta} < \boldsymbol{0}\\
\boldsymbol{0}&, \boldsymbol{\Theta}=\boldsymbol{0}\\
\boldsymbol{1}&, \boldsymbol{\Theta} > \boldsymbol{0}
\end{aligned}
\right. .
\end{equation}

Now, we can calculate the piecewise derivative for the objective function. By applying normal optimizer, we have a learnable model named ``{\gnlasso}-PW`` in this paper.

\subsection{Proximal Gradient Method}

Another way to learn the objective function with the $L_1$-norm based regularization term is to utilize the proximal method~\cite{parikh2014proximal}. Proximal method can optimize an objective function with a differentiable, Lipschitz-continuous part and a non-differentiable part. 

By given the proximal mapping for the lasso objective as follows:
\begin{equation}
	\begin{aligned}
\operatorname{prox}_{t}(\hat{\mb{\Theta}}) 
&=\arg \min _{\boldsymbol{z}} \frac{1}{2}\|\hat{\mb{\Theta}}-\boldsymbol{z}\|_{2}^{2}+\beta t\|\boldsymbol{z}\|_{1} \\
&=\mathcal{S}_{\beta t}(\hat{\mb{\Theta}}),\\
&\text{where } \left[\mathcal{S}_{\beta t}(\hat{\mb{\Theta}})\right]_{i}=\left\{\begin{array}{ll}
\hat{\mb{\Theta}}_i-\beta t, & \text { if } \hat{\mb{\Theta}}_i>\beta t \\
0, & \text { if }\left|\hat{\mb{\Theta}}_i\right| \leq \beta t \\
\hat{\mb{\Theta}}_i+\beta t, & \text { if } \hat{\mb{\Theta}}_i<-\beta t
\end{array}\right.,\\
&\qquad \quad \hat{\mb{\Theta}} = \mb{\Theta}- t \nabla \ell(\mb{\Theta}),
\end{aligned}
\end{equation}
we obtain the following proximal gradient update:
\begin{equation}
	\mb{\Theta}^{(k)}=\mathcal{S}_{\beta t}\left(\mb{\Theta}^{(k-1)}- t \nabla \ell(\mb{\Theta}^{(k-1)})\right) .
\end{equation}

To apply this method, we also need to prove the convergence. For the loss function $L = \ell(\mb{\Theta}) + \beta \cdot \left\| \mb{\Theta} \right\|_1$, we have the assumption that $g$ is convex, differentiable, and $\nabla g$ is Lipschitz continuous with $L>0$. Also, $\left\| \mb{\Theta} \right\|_1$ is convex and its proximal map can be evaluated. Then, we have the following Theorem:

\begin{theo}
Proximal gradient method with fixed step size~~~ $t<1/L$~~~ satisfies the following convergence rate:

\begin{equation}
L(\mb{\Theta}^{(k)})-L(\mb{\Theta}^{\star}) \leq \frac{\|\mb{\Theta}^{(0)}-\mb{\Theta}^{\star} \|^2_2}{2tk}.
\end{equation}

\end{theo}

\subsection{Accelerated Proximal Gradient Method}

Accelerated proximal gradient method is proposed to speed up the previous one. It's similar to the regular proximal gradient descent, but the argument passed to the $\operatorname{prox}$ has changed. It works as follows:

First, we choose the initial $\mb{\Theta}^{(0)}=\mb{\Theta}^{(-1)} \in \mathbb{R}^{n}$. The recursive formula for $\mb{\Theta}$ can be expressed as follows:

\begin{equation}
\begin{aligned}
v &= \mb{\Theta}^{(k-1)} + \frac{k-2}{k+1}(\mb{\Theta}^{(k-1)} - \mb{\Theta}^{(k-2)})\\
\mb{\Theta}^{(k)} &= \operatorname{prox}_t(v - t \nabla \ell(v))
\end{aligned}.
\end{equation}	

Also, the accelerated proximal gradient method is convergent here. With the same assumption, we have the following theorem:

\begin{theo}
Accelerated proximal gradient method with fixed step size $t<1/L$ satisfies the following convergence rate:

\begin{equation}
L(\mb{\Theta}^{(k)})-L(\mb{\Theta}^{\star}) \leq 2\frac{\|\mb{\Theta}^{(0)}-\mb{\Theta}^{\star} \|^2_2}{t(k+1)^2}.
\end{equation}
\end{theo}

%-----------------------------------------------
\section{Experiments}\label{sec:experiment}

To verify the effectiveness of the proposed method, we do the experiments on two real-world datasets, METR-LA~\cite{DBLP:conf/iclr/LiYS018} and Nasdaq-100~\cite{stock}. METR-LA contains the traffic speed data with 207 sensors and 1515 edges within 4 months, and Nasdaq-100 contains 96 stocks (without the ones that have missing values) and 182 edges within two years . In this section, we will fisrt introduce our experimental settings in section~\ref{setting}. Then, we show the main result in section~\ref{result}, and discuss the framework, gradient methods, parameters of the model in section~\ref{result}-\ref{beta}, and last, the case study in section~\ref{case}

\begin{table*}[h]
\caption{Performance of {\gnlasso} and other baselines on Nasdaq-100 and METR-LA dataset. Numbers with bolded font represent the best results, and numbers with underline are ranked second. From the table, {\gnlasso} achieves competitive results comparing to other baselines.}
\begin{tabular}{c|c|ccc|ccc|ccc}
\hline
\multirow{2}{*}{Data}        & \multirow{2}{*}{Method} & \multicolumn{3}{c|}{1 day}                          & \multicolumn{3}{c|}{3 days}                         & \multicolumn{3}{c}{5 days}                          \\
&                         & $\operatorname{MAE}$             & $\operatorname{RMSE}$            & $R^2$              & $\operatorname{MAE}$             & $\operatorname{RMSE}$            & $R^2$              & $\operatorname{MAE}$             & $\operatorname{RMSE}$            & $R^2$              \\ \hline
\multirow{10}{*}{Nasdaq-100} & Linear                  & 0.1370          & 0.2703          & 0.9527          & 0.1603          & 0.2776          & 0.9499          & 0.1830          & 0.3067          & 0.9375          \\
& GL                      & 0.1397          & 0.2696          & 0.9529          & 0.1686          & 0.2827          & 0.9480          & 0.1876          & 0.3117          & 0.9355          \\
& TVGL                    & 0.1585          & 0.2715          & 0.9517          & 0.1790          & 0.2872          & 0.9481          & 0.2172          & 0.3369          & 0.9295          \\
& LSTM                    & 0.1456          & 0.2704          & 0.9526          & 0.1655          & 0.2785          & 0.9496          & 0.1863          & 0.3080          & 0.9370          \\
& GCN                     & \underline{0.1321}          & \underline{0.2629}          & \underline{0.9552}          & \textbf{0.1512} & \textbf{0.2699}          & \textbf{0.9527}          & 0.1777          & 0.3024          & 0.9394          \\
& GAT                     & 0.1357          & 0.2641          & 0.9548          & 0.1566          & 0.2745          & 0.9510          & 0.1837          & 0.3045          & 0.9385          \\
& GWN                     & 0.1405          & 0.2717          & 0.9522          & 0.1620          & 0.2775          & 0.9500          & 0.1831          & 0.3066          & 0.9375          \\
& GMAN                    & 0.1409          & 0.2714          & 0.9523          & 0.1624          & 0.2805          & 0.9490          & 0.1803          & 0.3070          & 0.9376          \\
& MTGNN                   & \textbf{0.1311} & 0.2640          & 0.9548          & \textbf{0.1512} & 0.2722          & 0.9519          & \textbf{0.1727} & \underline{0.3008}          & \underline{0.9399}          \\
& {\gnlasso}                     & \underline{0.1321}          & \textbf{0.2618} & \textbf{0.9556} & 0.1514         & \underline{0.2706} & \underline{0.9525} & \underline{0.1739}          & \textbf{0.3000} & \textbf{0.9402} \\ \hline
\multirow{2}{*}{Data}        & \multirow{2}{*}{Method} & \multicolumn{3}{c|}{15 min}                         & \multicolumn{3}{c|}{30 min}                         & \multicolumn{3}{c}{60 min}                          \\
&                         & $\operatorname{MAE}$             & $\operatorname{RMSE}$            & $R^2$              & $\operatorname{MAE}$             & $\operatorname{RMSE}$            & $R^2$              & $\operatorname{MAE}$             & $\operatorname{RMSE}$            & $R^2$              \\ \hline
\multirow{10}{*}{METR-LA}     & linear                  & 3.4799          & 6.8002          & 0.7998          & 4.1158          & 7.9611          & 0.7226          & 5.1453          & 9.3595          & 0.6146          \\
& GL                      & 3.4418          & 6.8501          & 0.7968          & 4.0467          & 8.0738          & 0.7147          & 5.0440          & 9.5516          & 0.5986          \\
& TVGL                    & 3.4382          & 6.8562          & 0.7965          & 4.0461          & 8.0749          & 0.7146          & 5.0363          & 9.5665          & 0.5974          \\
& LSTM                    & 3.3677          & 6.6996          & 0.8057          & 3.9620          & 7.9363          & 0.7243          & 4.8860          & 9.3258          & 0.6174          \\
& GCN                     & 3.5620          & 6.6376          & 0.8093          & 4.1745          & 8.2989          & 0.6986          & 5.1828          & 9.3630          & 0.6143          \\
& GAT                     & 3.3806          & 6.8266          & 0.7982          & 4.0357          & 8.0825          & 0.7141          & 5.1660          & 9.6787          & 0.5879          \\
& GWN                     & \textbf{2.6900} & \underline{5.1500}          & \underline{0.8852}          & \underline{3.0700}          & 6.2200          & 0.8307          & 3.5300          & 7.3700          & 0.7610          \\
& GMAN                    & 2.7700          & 5.4800          & 0.8700          & \underline{3.0700}          & 6.3400          & 0.8241          & \textbf{3.4000} & \underline{7.2100 }         & \underline{0.7713}          \\
& MTGNN                   & \textbf{2.6900}          & 5.1800          & 0.8838          & \textbf{3.0500} & \underline{6.1700}          & \underline{0.8334}          & \underline{3.4900}          & 7.2300          & 0.7700          \\
& {\gnlasso}                     & 2.7928          & \textbf{5.1095} & \textbf{0.8869} & 3.0750          & \textbf{5.9493} & \textbf{0.8449} & 3.5972          & \textbf{7.1134} & \textbf{0.7765}\\
\hline
\end{tabular}
\label{tab:main}
\end{table*}

\subsection{Experimental Settings} \label{setting}

\subsubsection{Comparison Methods}

\hfill\\

To evaluate the proposed method, we compare {\gnlasso} with the following baseline methods.

\begin{itemize}
    \item \textbf{\linear}: We build the auto-regressive model for every entity separately~\cite{seber2012linear}. This model ignores the dependency among entities.
    \item \textbf{\gl}: We build linear model using all entities with graphical lasso to calculate the relationships between the entities~\cite{Friedman_Sparse_Bioinformatics_08}.
    \item \textbf{\tvgl}: Similar to {\gl} but use {\tvgl} to estimate the relationships~\cite{Hallac_Network_KDD_17}. {\tvgl} can learn the time-varying correlation but can't learn when the correlation changed.
    \item \textbf{\lstm}: {\lstm} is used to model every entity seperately~\cite{Hochreiter_Long_Neural_97}. It models a long sequence well but also misses the network information.
    \item \textbf{\gcn}: First introduced to solve the semi-supervised classification problem. We extend this method to solve the regression problem as regarding the network unchangeable over time~\cite{Kipf_Semi_CORR_16}. We consider the graphs at different time step share the same parameters and use the output without softmax. 
    \item \textbf{\gat}: Similar to {\gcn} but use attention mechanism to do aggregation~\cite{Velickovic_Graph_ICLR_18}. 
    \item \textbf{\gman}: A graph multi-attention network with spatial-temporal attentions and spatial-temporal embeddings~\cite{DBLP:conf/aaai/ZhengFW020}.
    \item \textbf{\gwn}: A spatial-temporal graph neural network with graph convolution layer and temporal convolution layer~\cite{wu2019graph}.
    \item \textbf{\mtgnn}: A graph neural network for multivariate time series data with graph structure learning, graph convolution and temporal convolution module~\cite{DBLP:journals/corr/abs-2005-11650}.
\end{itemize}

\subsubsection{Experimental Setups}

\hfill\\

\textbf{Datasets}: For METR-LA, we follow the data preprocessing as in~\cite{DBLP:conf/iclr/LiYS018}, and for Nasdaq-100, we do the similar process. We apply the z-score normalization on the original dataset and split it with 70\% for training, 10\% for validation and 20\% for testing. Then we set the window size as 5 to predict the next 1, 3 and 5 days. 

\textbf{Hyperparameters}: We do the experiments with one {\gnlasso} layer, one LeakyRelu layer and a fully connected prediction layer. We train our model using Adam optimizer~\cite{DBLP:journals/corr/KingmaB14} with the learning rate of 0.001. Also, we randomly initialize the input hidden state vector using the normal distribution. Last of all, there are four hyperparameters in our model, i.e., the hidden size, the aggregation size, the coefficient of $L_1$-norm $\beta$, and the negative slope of LeakyRelu. For METR-LA, we set the hidden size to 32, the aggregation size to 16, $\beta$ to 0.00002 and the negative slope to 0.5. For Nasdaq-100, we set the hidden size to 64, aggregation size to 32, $\beta$ to 0.002 and the negative slope to 0.5.

\textbf{Metrics}: In our experiments, we use three evaluation metrics, i.e., Mean Absolute Error (MAE), Root Squared Error (RMSE), and $R^2$.

\subsection{Experimental Results} \label{result}

Table~\ref{tab:main} shows the results of {\gnlasso} and other baseline methods on Nasdaq-100 and METR-LA datasets. It is observed that {\gnlasso} is competitive with other baseline methods on both datasets. {\gnlasso} achieves better results in two of the three metrics. In the following, we'll discuss the results of these two datasets. Due to the different characteristics of the data, these methods show different results.

For stock data, it's hard to determine the relationships since two stocks may have competition, co-operation, or no relationship. Also, the stock data is sensitive to the different news in the market. The adjacency matrix we used here is the correlation matrix calculated by {\gl}. It just gives a brief view of the graph structure. Thus, if the model can't adaptively learn the adjacency matrix well, it may not get an ideal result. {\gnlasso} gets the best results over
other baseline methods on $\operatorname{RMSE}$ and $\operatorname{R^2}$. Although, the improvement is limited. We believe it's possibly due to the reason that the relationships between the stock change dramatically over time. Much complicated model may not learn the result better.

As for the METR-LA dataset, the entities are highly correlated by their location. And it's obvious that the relationships will change over time. Also, The traffic conditions will be different at different times of the day. But this change is much stable than the stock data. From Table~\ref{tab:main}, {\gnlasso} achieves much more significant improvements. Besides, combining sequence and graph information does improve accuracy on this dataset. {\gwn}, {\gman}, {\mtgnn}, {\gnlasso} work better than the methods that only use the sequence information or the graph information. Also, those GNNs which doesn't maintain RNN cell, like GWN, may be faster but fail to capture the information in sequence precisely.

\begin{table*}[h]
\caption{Performance on four variants of {\gnlasso}. Without the attention mechanism or the $L_1$-norm regularization, the model performs worse.}
\begin{tabular}{c|l|ccc|ccc|ccc}
\hline
\multirow{2}{*}{Data}       & \multicolumn{1}{c|}{\multirow{2}{*}{Method}} & \multicolumn{3}{c|}{1 day}                          & \multicolumn{3}{c|}{3 days}                         & \multicolumn{3}{c}{5 days}                          \\
& \multicolumn{1}{c|}{}                        & $\operatorname{MAE}$             & $\operatorname{RMSE}$            & $R^2$              & $\operatorname{MAE}$             & $\operatorname{RMSE}$            & $R^2$              & $\operatorname{MAE}$             & $\operatorname{RMSE}$            & $R^2$              \\ \hline
\multirow{4}{*}{Nasdaq-100} & {\gnlasso}                                          & \textbf{0.1321} & \textbf{0.2618} & \textbf{0.9556} & \textbf{0.1514} & \textbf{0.2706} & \textbf{0.9525} & \textbf{0.1739} & \textbf{0.3000} & \textbf{0.9402} \\
& {\gnlasso}-NL                                 & 0.1513          & 0.2791          & 0.9495          & 0.1677          & 0.2855          & 0.9471          & 0.1840          & 0.3128          & 0.9352          \\
& {\gnlasso}-NG     &0.1392     &0.2700  &0.9528  &0.1616  &0.2794  &0.9494  &0.1828  &0.3060  &0.9380                    \\
& {\gnlasso}-Fix  &0.1438   &0.2717  &0.9522  &0.1611  &0.2781  &0.9498  &0.1808  &0.3056  &0.9381          \\ \hline
\multirow{2}{*}{Data}       & \multicolumn{1}{c|}{\multirow{2}{*}{Method}} & \multicolumn{3}{c|}{15 min}                         & \multicolumn{3}{c|}{30 min}                         & \multicolumn{3}{c}{60 min}                          \\
& \multicolumn{1}{c|}{}                        & $\operatorname{MAE}$             & $\operatorname{RMSE}$            & $R^2$              & $\operatorname{MAE}$             & $\operatorname{RMSE}$            & $R^2$              & $\operatorname{MAE}$             & $\operatorname{RMSE}$            & $R^2$              \\ \hline
\multirow{4}{*}{METR-LA}    & {\gnlasso}                                          & 2.7928          & \textbf{5.1095} & \textbf{0.8869} & \textbf{3.0750} & \textbf{5.9493} & \textbf{0.8449} & 3.5972          & \textbf{7.1134} & \textbf{0.7765} \\
& {\gnlasso}-NL                                 & 2.7735          & 5.1419          & 0.8854          & 3.0831          & 5.9536          & 0.8447          & 3.6207          & 7.1786          & 0.7725          \\
& {\gnlasso}-NG                                 & 2.8076          & 5.2549          & 0.8803          & 3.1046          & 6.0543          & 0.8392          & 3.6435          & 7.2141          & 0.7700          \\
& {\gnlasso}-Fix                              & \textbf{2.7830} & 5.1750          & 0.8839          & 3.0818          & 6.0005          & 0.8422          & \textbf{3.5671} & 7.1271          & 0.7756    \\ \hline     
\end{tabular}
\label{tab:gnl}
\end{table*}

\subsection{Ablation Study} \label{ablation}

To investigate the effectiveness of each part on the framework of {\gnlasso} , we do experiments with several variations of {\gnlasso}. Here, we test two main parts of {\gnlasso}. First, we do the experiment without $L_1$-norm to test the effectiveness of Lasso, named {\gnlasso}-NL in Table~\ref{tab:gnl}. Second, we modify the aggregation part to verify the necessity of attention mechanism. For the first one ({\gnlasso}-NG), we just throw away the aggregation operator and use the hidden state vector as input twice. For the second one ({\gnlasso}-Fix), instead of using the weighted average of hidden states, we use the mean of hidden states.

Table~\ref{tab:gnl} shows the results of ablation study. Due to the complexity of the relationships between stocks, removing the aggregation part or using a fixed graph doesn't lead to too bad results. But the lasso part actually improves the result. Meanwhile, the graph structure is much more evident in the traffic dataset. Without the aggregation part, {\gnlasso} gets the worst result. And with the fixed initialized adjacency matrix, the model is still competitive. Also, lasso improves the results for METR-LA. So far, we validate the effectiveness of the main parts of {\gnlasso}.

\subsection{Discussion of Different Gradient Methods} \label{gradient}

\begin{table*}
\caption{The result with different gradient methods}
\begin{tabular}{c|l|lll|lll|lll}
\hline
\multirow{2}{*}{Data}       & \multicolumn{1}{c|}{\multirow{2}{*}{Method}} & \multicolumn{3}{c|}{1 day}  & \multicolumn{3}{c|}{3 days} & \multicolumn{3}{c}{5 days} \\
                            & \multicolumn{1}{c|}{}                        & MAE      & RMSE    & R2     & MAE     & RMSE    & R2      & MAE     & RMSE    & R2     \\ \hline
\multirow{4}{*}{Nasdaq-100} & {\gnlasso}-F & 0.1443   & 0.2732  & 0.9516 & 0.1650  & 0.2821  & 0.9484  & 0.1839  & 0.3073  & 0.9376 \\
                            & {\gnlasso}-PW                                       & \textbf{0.1321}   & \textbf{0.2618}  & \textbf{0.9556} & \textbf{0.1514}  & \textbf{0.2706}  & \textbf{0.9525}  & \textbf{0.1739}  & \textbf{0.3000}  & \textbf{0.9402} \\
                            & {\gnlasso}-PG                                       & 0.1425   & 0.2705  & 0.9526 & 0.1687  & 0.2852  & 0.9471  & 0.1875  & 0.3134  & 0.9349 \\
                            & {\gnlasso}-APG                                      & 0.1346   & 0.2656  & 0.9543 & 0.1620  & 0.2774  & 0.9500  & 0.1821  & 0.3053  & 0.9382 \\ \hline
\multirow{2}{*}{Data}       & \multicolumn{1}{c|}{\multirow{2}{*}{Method}} & \multicolumn{3}{c|}{15 min} & \multicolumn{3}{c|}{30 min} & \multicolumn{3}{c}{60 min} \\
                            & \multicolumn{1}{c|}{}                        & MAE      & RMSE    & R2     & MAE     & RMSE    & R2      & MAE     & RMSE    & R2     \\ \hline
\multirow{4}{*}{METR-LA}    & {\gnlasso}-F                                       & 2.8181   & 5.2411  & 0.8808 & 3.1314  & 6.0140  & 0.8414  & 3.7432  & 7.2530  & 0.7676 \\
                            & {\gnlasso}-PW                                       & \textbf{2.7928}   & \textbf{5.1095}  & \textbf{0.8869} & \textbf{3.0750}  & \textbf{5.9493}  & \textbf{0.8449}  & \textbf{3.5972}  & \textbf{7.1134}  & \textbf{0.7765} \\
                            & {\gnlasso}-PG                                       & 2.7992   & 5.1864  & 0.8833 & 3.1510  & 6.1642  & 0.8330  & 3.6220  & 7.1668  & 0.7731 \\
                            & {\gnlasso}-APG                                      & 2.8635   & 5.1587  & 0.8847 & 3.1681  & 6.0578  & 0.8392  & 3.6311  & 7.1429  & 0.7746\\ \hline
\end{tabular}
\label{tab:gradient}
\end{table*}

\begin{figure}
    \begin{subfigure}[b]{0.45\columnwidth}
       \includegraphics[width=\linewidth]{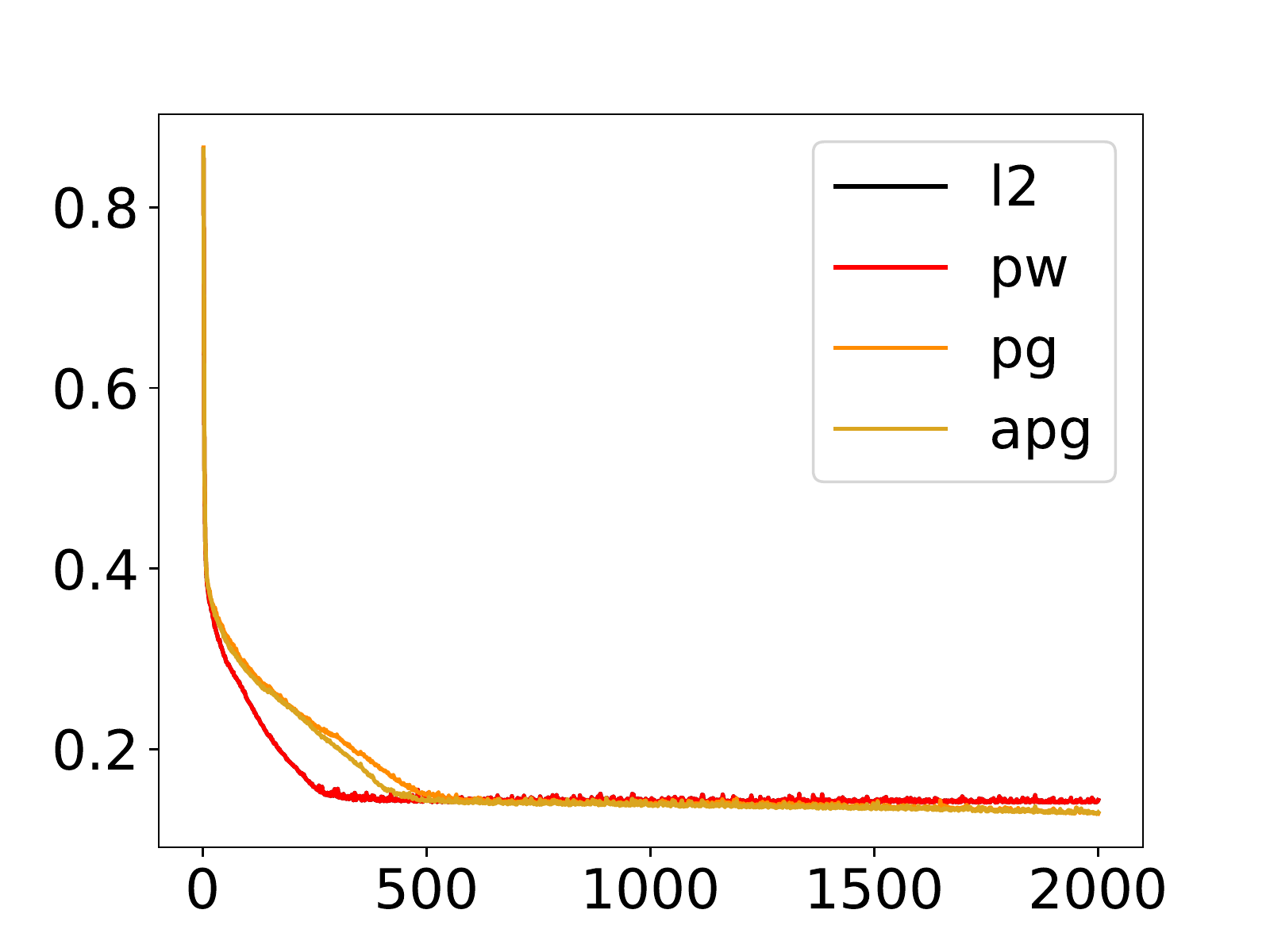}
       \caption{Loss on Nasdaq-100}
    \end{subfigure}  
    \begin{subfigure}[b]{0.45\columnwidth}
       \includegraphics[width=\linewidth]{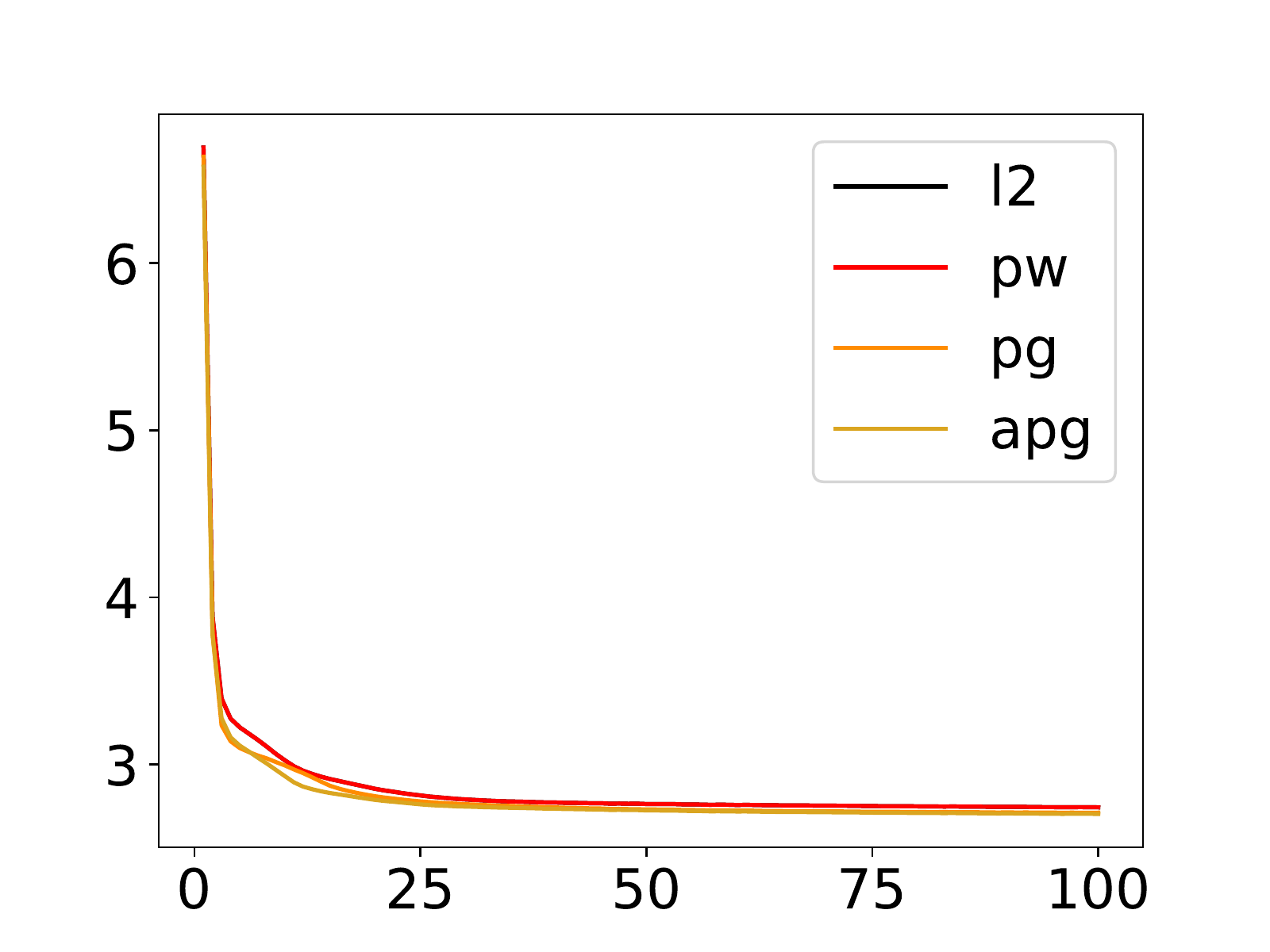}
       \caption{Loss on METR-LA}
    \end{subfigure}
    \caption{The training loss with different gradient method. Those four gradient methods all converge well on two datasets.}
    \label{fig:loss}
\end{figure}

Here, we show the results with different derivation methods of $L_1$-norm and the loss curve to prove its efficiency. Table~\ref{tab:gradient} shows the results with different learning techniques. We do the experiments with F-norm ({\gnlasso}-F), piecewise derivative ({\gnlasso}-PW), proximal gradient method ({\gnlasso}-PG) and accelerated proximal gradient method ({\gnlasso}-APG). And figure~\ref{fig:loss} shows the training loss curve of these four methods. It's obvious that our model converges well. At the beginning of the training, the loss value drops greatly, indicating that the learning rate is appropriate and the gradient descent process works well. After learning to a certain stage, the loss curve tends to be stable.

\subsection{Parameter Study on $\beta$} \label{beta}

\begin{figure}
    \begin{subfigure}[t]{0.45\columnwidth}
       \includegraphics[width=\linewidth]{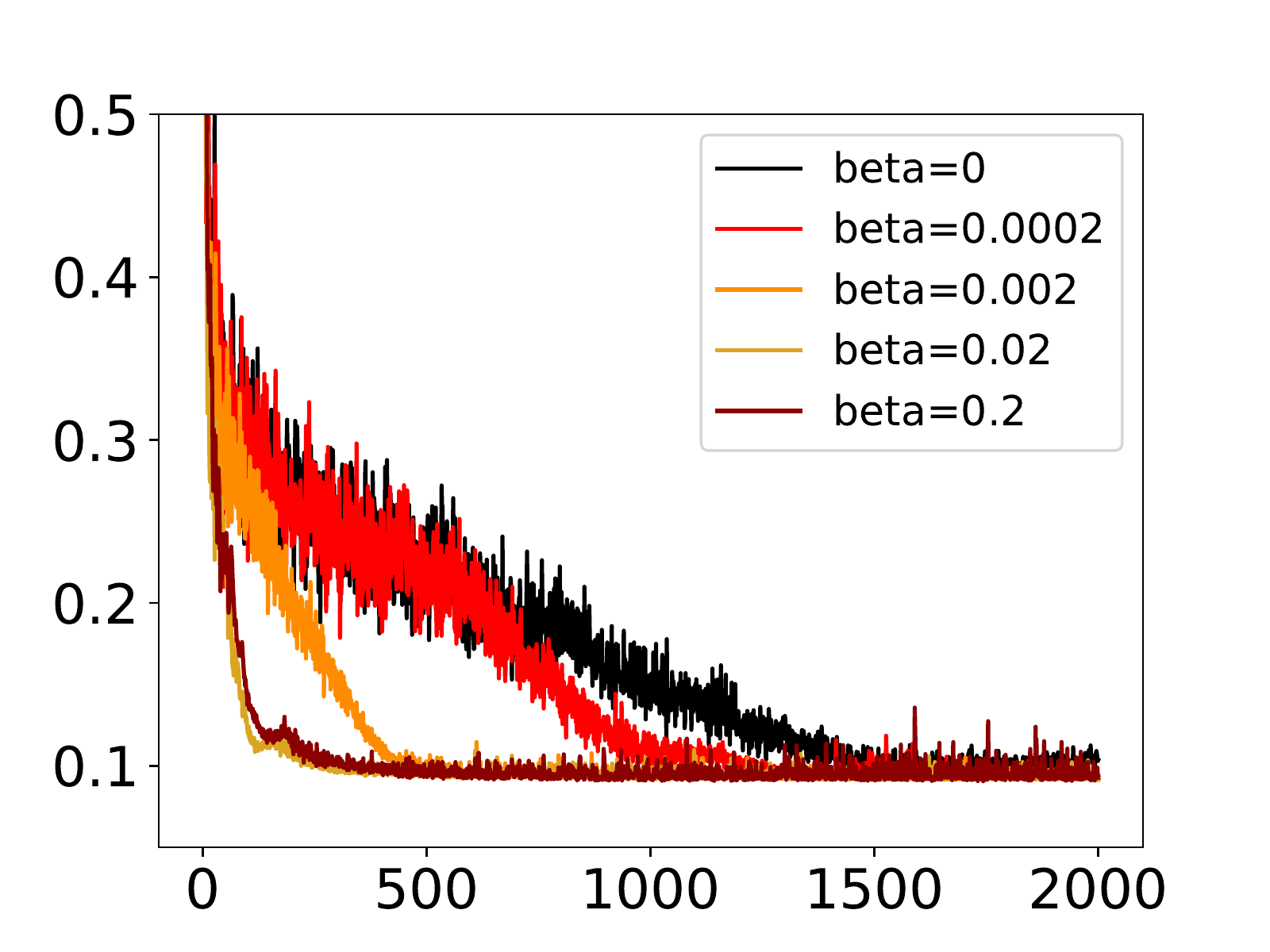}
       \caption{Loss on Nasdaq-100}
    \end{subfigure}
    \begin{subfigure}[t]{0.45\columnwidth}
       \includegraphics[width=\linewidth]{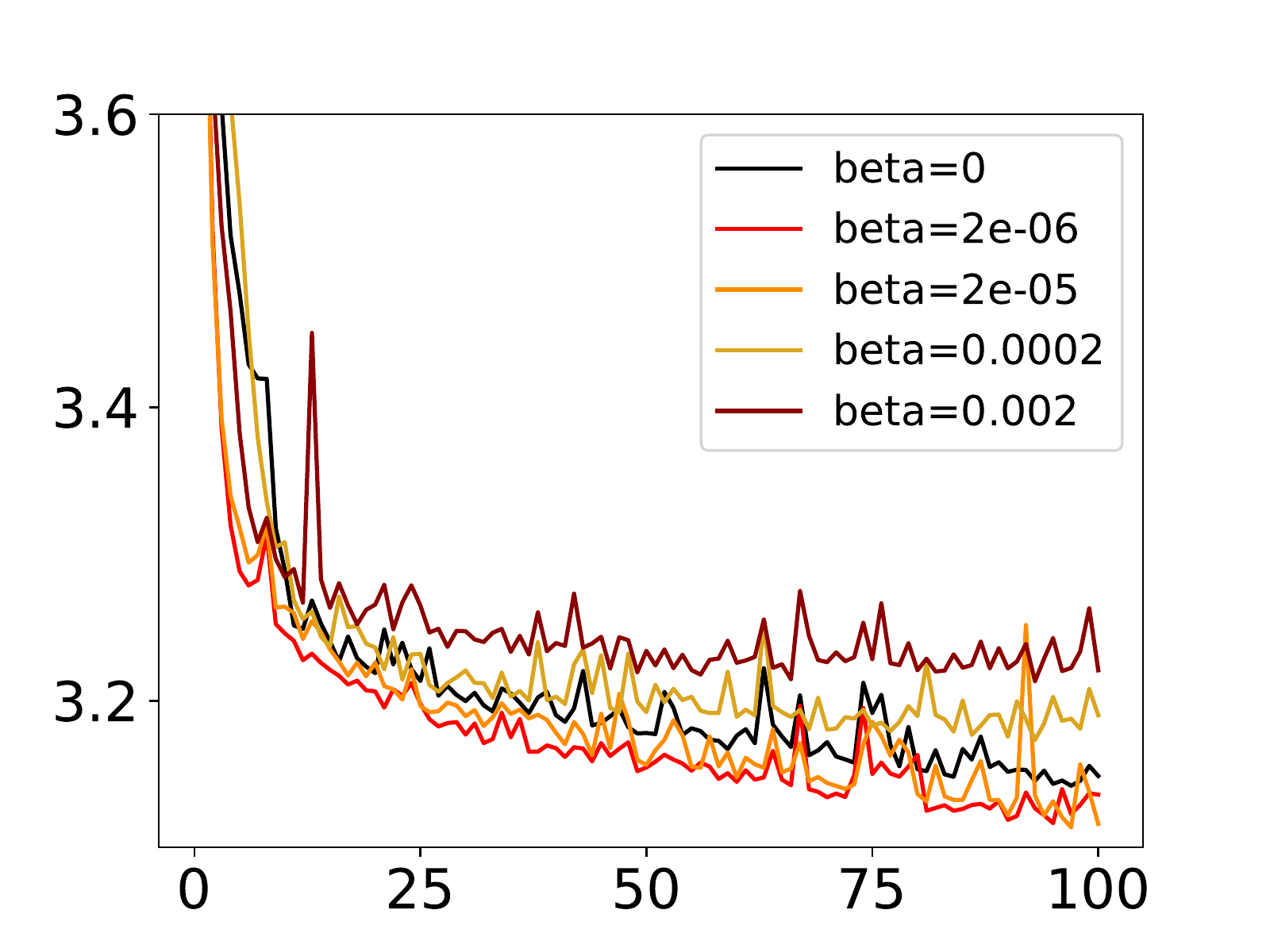}
       \caption{Loss on METR-LA}
    \end{subfigure}
    \caption{The test loss with different value of $\beta$. Adding $L_1$-norm helps model converge faster and avoid overfitting.}
    \label{fig:beta}
\end{figure}

To validate the effectiveness of $L_1$-norm term, we also do the experiments with different values of $\beta$ on the two datasets. The figure~\ref{fig:beta} shows the test loss curves with {\gnlasso}-PW. It's obvious that, with a different $\beta$, the test loss curve conveys different characteristics. Almost all the models with $L_1$-norm term converges faster than the model without $L_1$-norm term. We believe that $L_1$-norm term can avoid the fluctuation of parameters which makes the model converge faster. Also, the result shows that the model with $L_1$-norm term gets a better result than the model without $L_1$-norm term. Accordingly, adding the $L_1$-norm term can help the model converge faster and avoid overfitting.

\subsection{Case study} \label{case}

\begin{figure}
\centering
    \begin{subfigure}[t]{0.9\columnwidth}
       \centering
       \includegraphics[width=1\columnwidth]{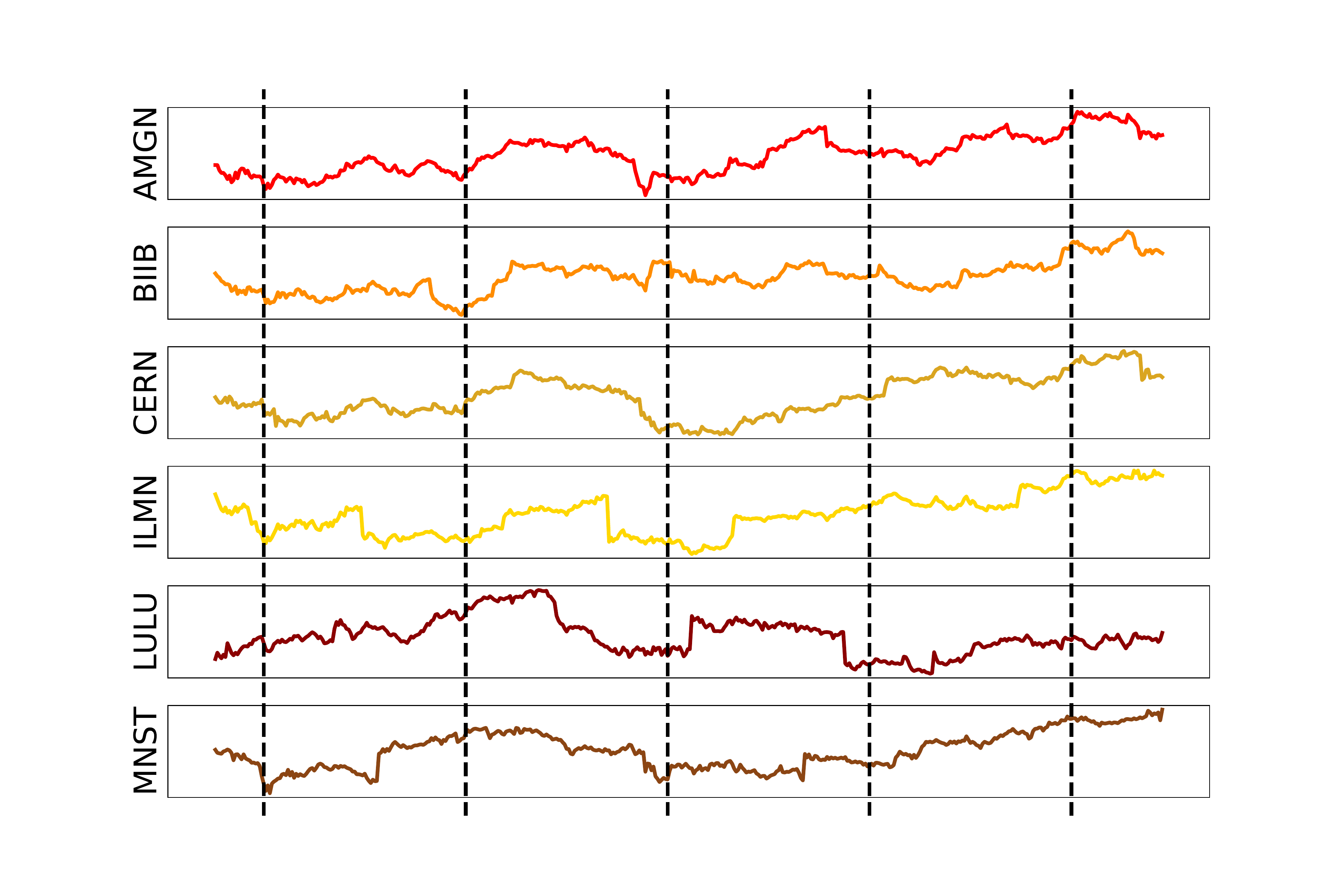}
    \end{subfigure}
 \hfill     

    \begin{subfigure}[t]{0.9\columnwidth}
       \centering
       \includegraphics[width=1\columnwidth]{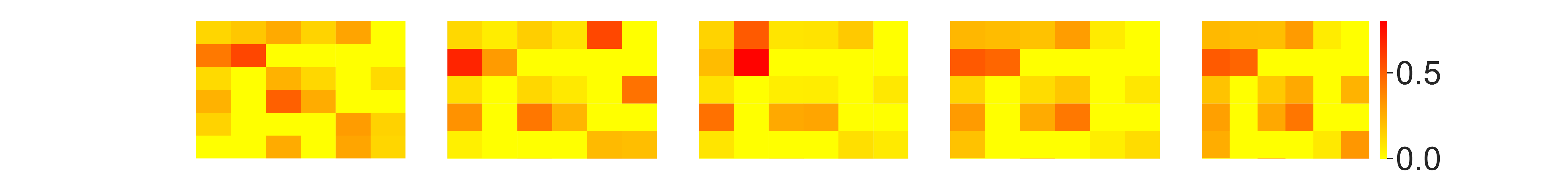}
    \end{subfigure}
    \caption{The biotechnology stock values and changing heatmaps of the learned attention. The heatmaps show high correlations between stocks.}
    \label{fig:case_stock}
\end{figure}

\begin{figure}
  \begin{subfigure}[b]{0.45\columnwidth}
    \includegraphics[width=\linewidth]{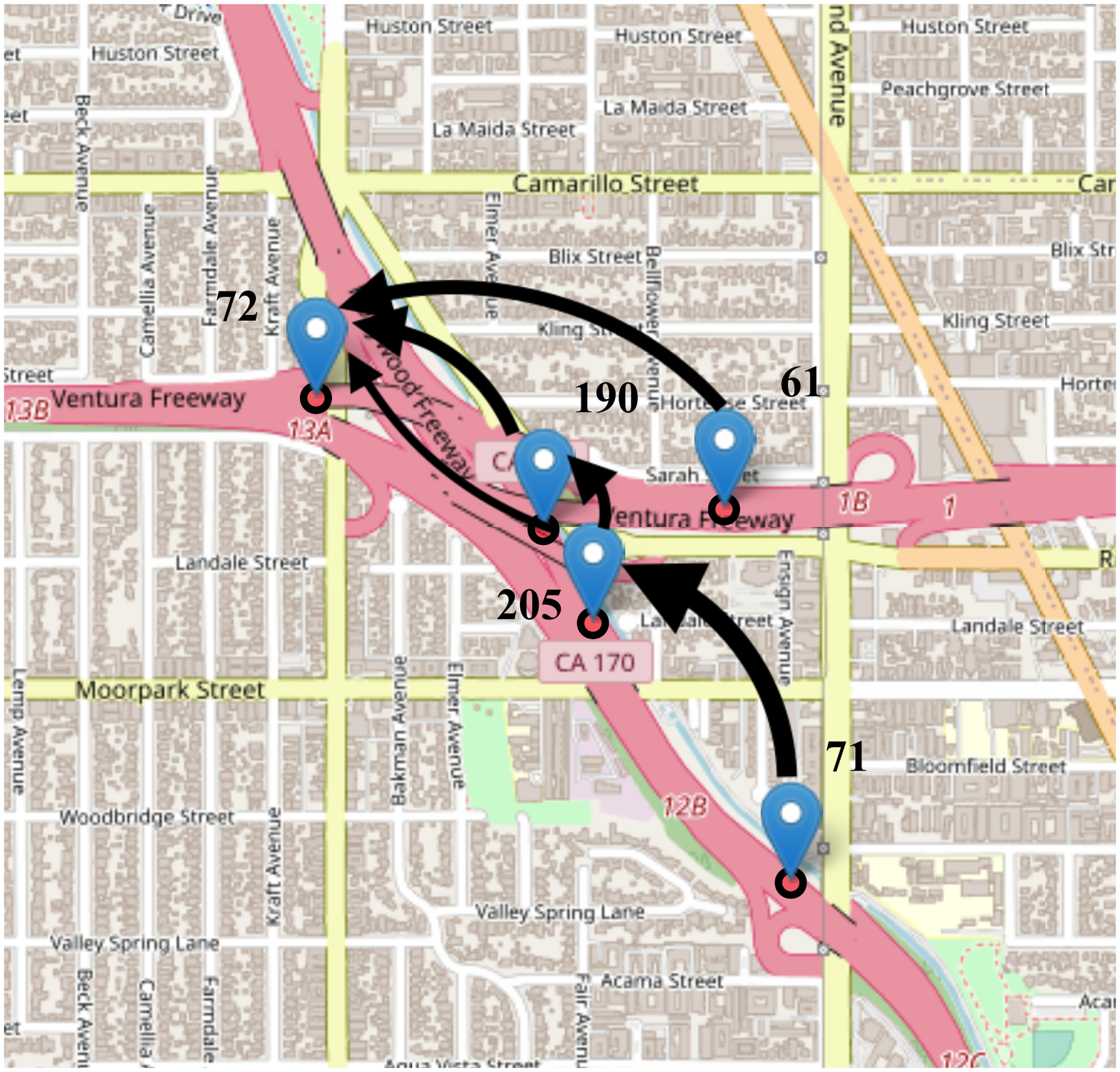}
    \caption{The geographical location of five sensors around a intersection in Los Angles.}
    \label{fig:case_traffic_a}
  \end{subfigure}
  % \hfill %%
  \begin{subfigure}[b]{0.54\columnwidth}
    \includegraphics[width=\linewidth]{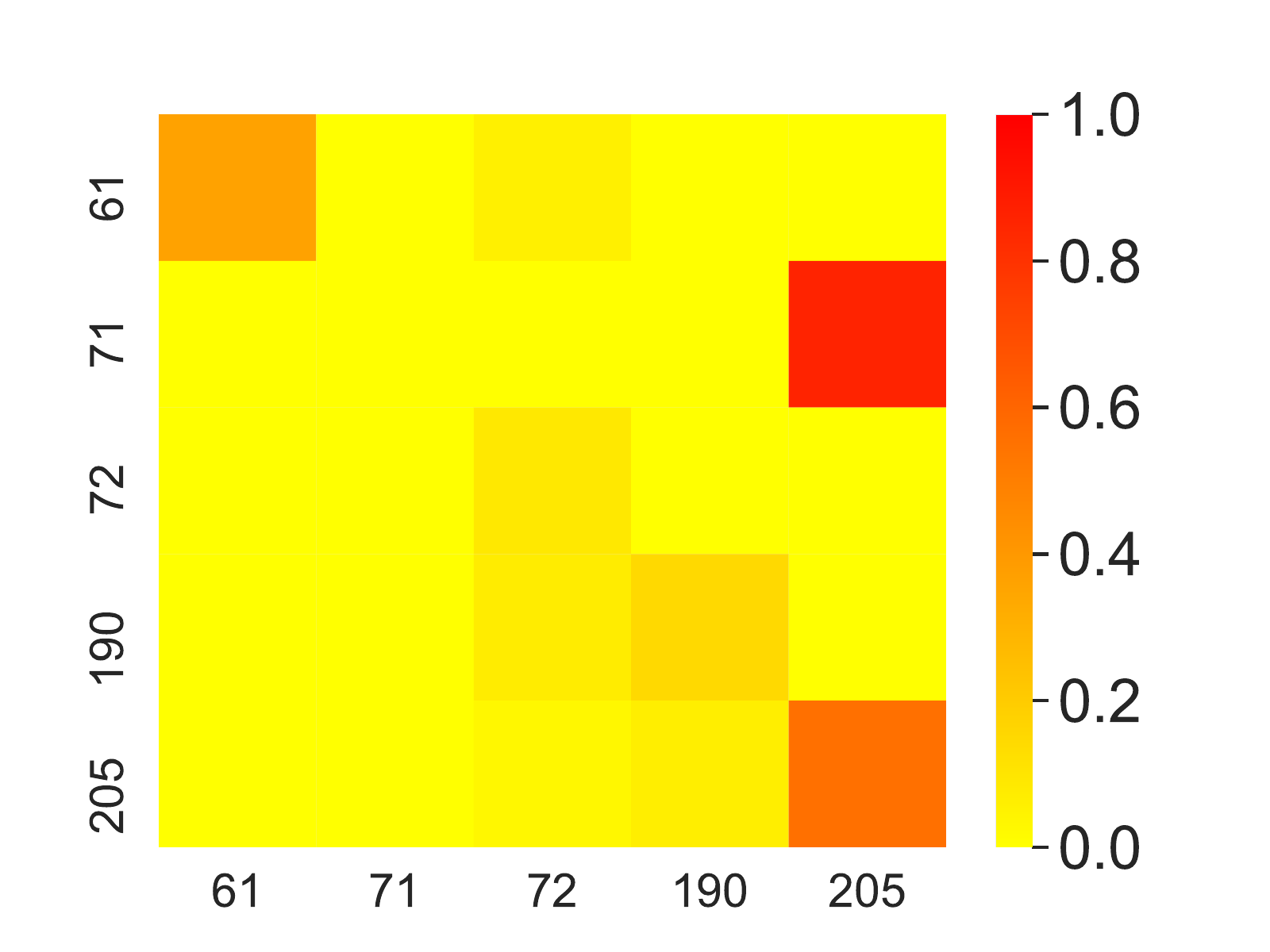}
    \caption{The heatmap of correlations between five sensor at 5 pm.}
    \label{fig:case_traffic_b}
  \end{subfigure}
  \caption{The learned attention for METR-LA.}
  \label{fig:case_traffic}
\end{figure}

To give a more direct sense of changing pattern in real-world applications, we show the learned heatmap in this part. We take six stocks of biotechnology companies as an example: AMGN, BIIB, CERN, ILMN, LULU and MNST. The figure~\ref{fig:case_stock} shows the fluctuation curve of these six stocks and the heatmap of the attentions at five time steps. It's obvious that the model successfully learns the highly correlated relationship between those companies. And the relationships are changing over time.

% \begin{itemize}
%   \item \textbf{1. AMGN:} Amgen Inc., American multinational biopharmaceutical company
%   \item \textbf{2. BIIB:} Biogen Inc., American multinational biotechnology company
%   \item \textbf{3. CERN:} Cerner Corporation, an American supplier of health information technology
%   \item \textbf{4. ILMN:} Illumina, Inc., American biotechnology company
%   \item \textbf{5. LULU:} Lululemon Athletica, athletic apparel retailer
%   \item \textbf{6. MNST:} Monster Beverage Corporation, American beverage company
% \end{itemize}

Another example is one intersection in Los Angeles. The figure~\ref{fig:case_traffic_a} shows five sensors near an intersection at the northwest of the downtown. And the figure~\ref{fig:case_traffic_b} is the heatmap of attentions learned at 5 pm. Then we use lines with arrows to plot the relationships. Since it's rush hour, the person leaving downtown should be more than entering downtown. The relationships learned from the model show a similar result here. We could find that sensor \#71 has a strong impact on sensor \#205.

%-----------------------------------------------

In this paper, we present a novel model for dynamic network regression. The basic cell of {\gnlasso} is based on {\gdu} which can capture the latent trends among the sequence. In addition, adding attentions to aggregate the information from related entities can learn the dynamic relationships between the entities. Our method combines the recurrent framework and the attention mechanism to better model the dynamic network problem. What's more, the $L_1$-norm regularization accelerates the learning procedure and avoids overfitting. On two public dynamic network datasets, METR-LA and Nasdaq-100, {\gnlasso} achieves competitive results with other baseline methods.

%-----------------------------------------------
\balance
\bibliographystyle{plain}
\bibliography{reference}

\end{document}